\documentclass[runningheads]{llncs}
\usepackage[T1]{fontenc}
\usepackage{graphicx,verbatim}
\usepackage{booktabs} 
\usepackage{siunitx} 
\usepackage{float} 
\usepackage{caption}
\usepackage{subcaption} 
\begin{document}
\title{HQColon: A Hybrid Interactive Machine Learning Pipeline for High Quality Colon Labeling and Segmentation} %
\author{Martina Finocchiaro \inst{1}\and
Ronja Stern\inst{1}\and
Abraham George Smith\inst{1,2}\and
Jens Petersen\inst{1,2}\and
Kenny Erleben\inst{1}\and
Melanie Ganz\inst{1,3}
}
\authorrunning{M. Finocchiaro et al.}
\titlerunning{HQColon}
%
\institute{Department of Computer Science, University of Copenhagen, Copenhagen, Denmark \and Department of Clinical Oncology, Center for Cancer and Organ Diseases, Copenhagen University Hospital, Copenhagen, Denmark\and Neurobiology Research Unit, Copenhagen University Hospital, Copenhagen, Denmark}

\maketitle              
\begin{abstract}
High-resolution colon segmentation is crucial for clinical and research applications, such as digital twins and personalized medicine. However, the leading open-source abdominal segmentation tool, \textit{TotalSegmentator}, struggles with accuracy for the colon, which has a complex and variable shape, requiring time-intensive labeling.
Here, we present the first fully automatic high-resolution colon segmentation method. To develop it, we first created a high resolution colon dataset using a pipeline that combines region growing with interactive machine learning to efficiently and accurately label the colon on CT colonography (CTC) images. Based on the generated dataset consisting of 435 labeled CTC images we trained an nnU-Net model for fully automatic colon segmentation. Our fully automatic model achieved an average symmetric surface distance of 0.2 mm (vs. 4.0 mm from \textit{TotalSegmentator}) and a 95th percentile Hausdorff distance of 1.0 mm (vs. 18 mm from \textit{TotalSegmentator}).
Our segmentation accuracy substantially surpasses \textit{TotalSegmentator}. We share our trained model and pipeline code, providing the first and only open-source tool for high-resolution colon segmentation. Additionally, we created a large-scale dataset of publicly available high-resolution colon labels.
\end{abstract}
\section{Introduction} 
\begin{figure}[tb]
    \centering    \includegraphics[width=0.9\linewidth]{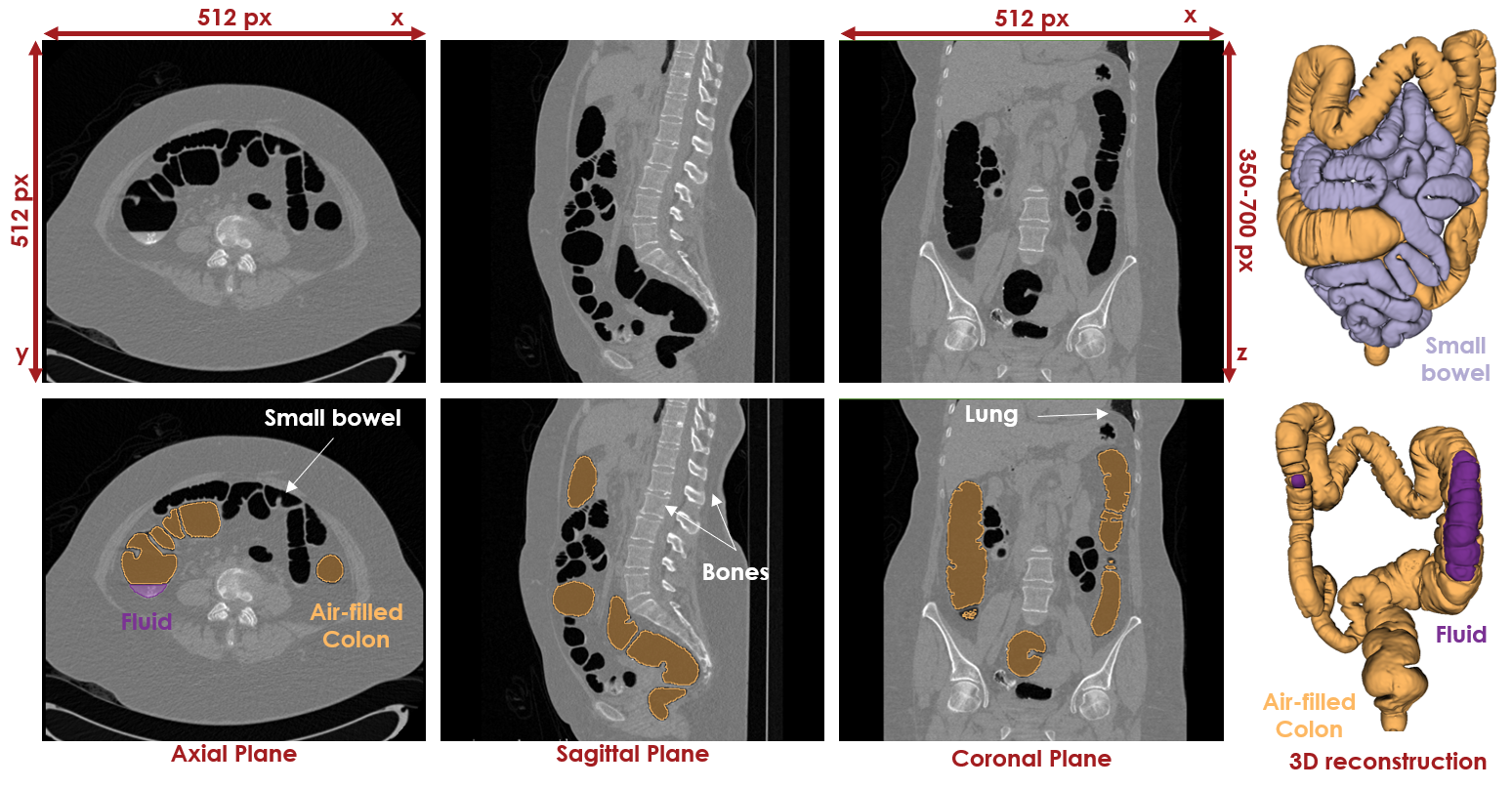}
    \caption{Example of axial, sagittal, and coronal CT colonography slices (top) with corresponding air- and fluid-filled colon annotations (bottom). On the right, 3D reconstructions show the colon alone (bottom) and with the small bowel (top).}
    \label{fig:ct_colonography}
\end{figure}

\begin{figure}[htbp]
    \centering    \includegraphics[width=0.99\linewidth]{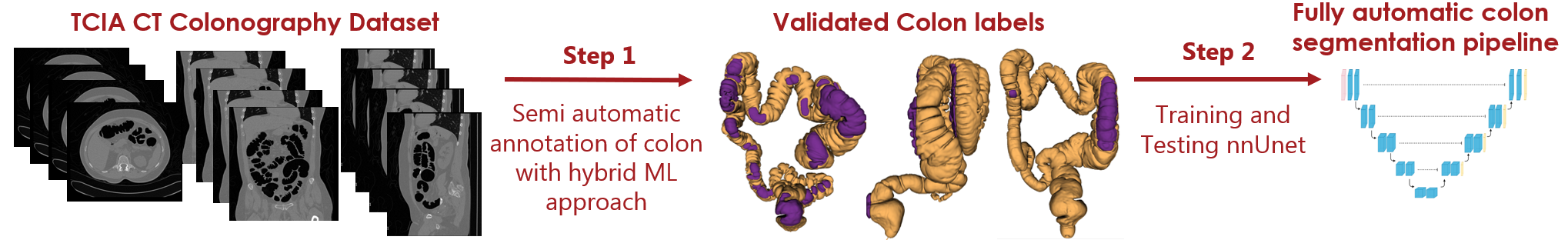}
    \caption{Two steps pipeline: 1) generation of high-resolution colon-labeled dataset 2) training and testing for fully automated colon segmentation.}
    \label{fig:method}
\end{figure}

CT colonography (CTC) is a non-invasive imaging technique to detect and monitor colorectal abnormalities. The procedure involves inflating the colon with air and capturing CT images ~\cite{mang2007ct}. Its effectiveness in both clinical applications and large-scale studies is dependent on accurate segmentation \cite{park2010computer},\cite{cirillo2019big}. Automating the segmentation process is essential to generate extensive organ datasets supporting the advancements of digital twins, \textit{in-vitro} clinical trials, and AI-driven personalized medicine \cite{tang2024roadmap}, \cite{meijer2023digital} \cite{liu2024towards}, \cite{vallee2024envisioning}.\\
Colon segmentation is challenging due to its flexible anatomy, surrounding air-filled structures, and variability in size, shape, and position. Collapsed sections and residual fluids add further complexity (Fig. \ref{fig:ct_colonography}). Traditional segmentation methods, using thresholding, region growing, and morphological operations, rely on heuristic rules that require manual feature selection and lack generalizability \cite{masutani2001automated}, \cite{chowdhury2011fast}, \cite{bert2009automatic}, \cite{iordanescu2005automated}, \cite{ismail2012fully}.
Deep learning approaches, particularly CNNs and U-Nets, enhance segmentation by reducing manual effort and handling anatomical variability  \cite{yu2022multi}, \cite{guachi2019automatic}. However, their effectiveness depends on large annotated datasets, which are time-consuming to produce for complex 3D organs like the colon. For instance, expert colon annotation with software like \textit{3D Slicer} \cite{kikinis20133d} takes 10–30 minutes per case, in our experience. Models like \textit{TotalSegmentator} show promise but generate coarse segmentations that fail when encountering  colon-specific challenges \cite{wasserthal2023totalsegmentator}.\\
We developed an open-source tool, \textit{HQColon}, for fully automatic high-resolution colon segmentation from CTC images. Traditional segmentation methods struggle with highly variable datasets, and manual annotation is too time-intensive. To address these issues, we: 1) developed a semi-automatic pipeline for colon annotation with minimal user input; 2) used the above approach to generate a training dataset with expert-validated labels (Fig. \ref{fig:method}, step 1) and 3) trained and tested a neural network for fully automatic colon segmentation (Fig. \ref{fig:method}, step 2).
The trained model and expert validated annotated dataset are publicly available [anonymous]. To the best of our knowledge, our presented method is the first open-source tool for high-resolution colon segmentation from CTC images.
\section{Methods} 
\subsection{Initial Dataset}
We used the publicly accessible CTC dataset from The Cancer Imaging Archive (TCIA) \cite{smith2015data}. The dataset includes 825 outpatients aged 50 and older, scheduled for colonoscopy screening with no procedures in the past five years. It consists of 3,451 CT scans with a spatial resolution around 0.8 mm. Scans were excluded if they exhibited dimensional inconsistencies, including (1) fewer than 350 or more than 700 axial slices, (2) axial slices smaller than 512×512 pixels, or (3) a disrupted format.
\subsection{Semi-automatic annotation of the air-filled colon segments} 
Filled with air, the colon appears darker in CT colonography compared to most of the surrounding organs (Fig. \ref{fig:ct_colonography}). To quickly generate high-quality colon annotations for training a fully automated segmentation model, we first applied traditional segmentation methods based on simple rules. The images were converted to binary format using a threshold of -800 HU. Fast annotation of air-filled colon regions was achieved by applying a 3D 26-neighbor region growing algorithm. The seed was placed by automatically extracting a region along the left-right midline, spanning ±50 pixels around the anterior-posterior midpoint, and selecting slices from index 50 to 250 along the inferior-superior axis. The first air-filled pixel encountered when scanning upward was chosen as the seed, ensuring placement in the rectum in most cases. This selection is based on three key observations: (1) patients are well-centered along the left-right axis but vary along the anterior-posterior axis, (2) the first air-filled region from an inferior-to-superior scan is likely the colon, and (3) some upper-body scans may include the proximal legs. Cases where automatic seed placement failed were excluded, as this phase aimed for fast colon annotation rather than a universally applicable method. To account for potential colon collapse or connections to other organs like the small bowel, segmentations with volumes exceeding 27 $cm^3$ or below 3.5 $cm^3$ were discarded. \\
The final colon annotations were validated by an expert, and incorrect ones were removed (\textit{e.g.,} with collapsed segments or incorrectly connected to other organs). The resulting annotated and validated images represent the dataset used in the second part of the study to train and test nnU-Net for fully automatic colon segmentation. 
\subsection{Interactive Machine Learning annotation of fluid-pockets} \label{subsec:fluid}
\begin{figure}[tb]
    \centering    \includegraphics[width=0.99\linewidth]{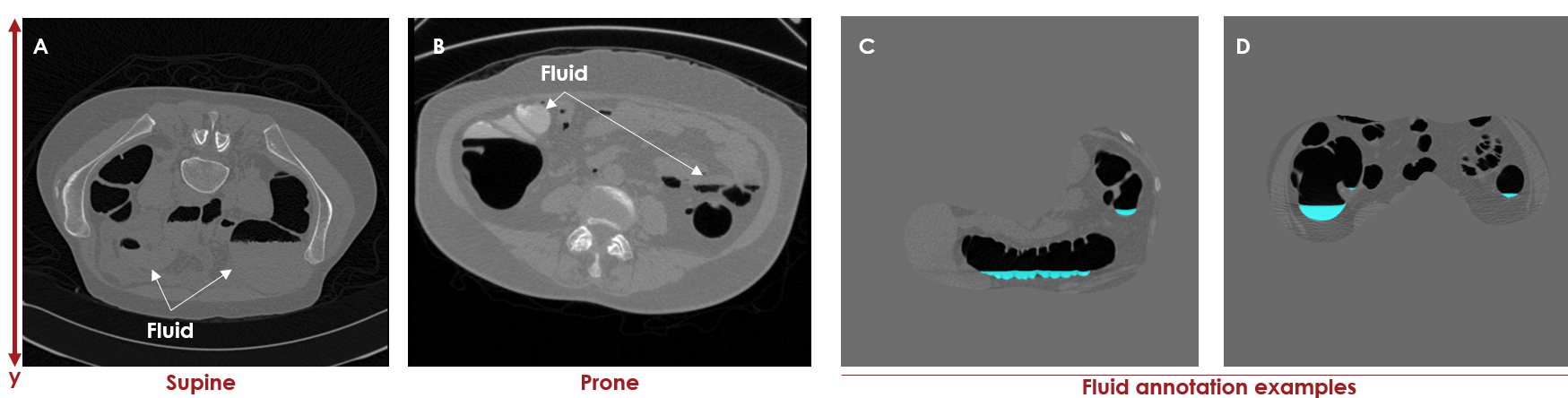}
    \caption{Examples of fluid visualization on axial slices: (A) supine and (B) prone patient. C, D show  \textit{RootPainter} fluid segmentation.}
    \label{fig:fluid}
\end{figure}

In CTC, the colon contains fecal residues along with air, which appear as fluid pockets in the images. This fluid varies in color, covering the full range of CTC pixel intensities, making traditional segmentation methods ineffective (Fig. \ref{fig:fluid}). To address this issue, we used an interactive machine learning approach with \textit{RootPainter} \cite{smith2022rootpainter}.\\
The dataset was prepared by generating colon masks using \textit{TotalSegmentator} on the raw images from a subset of the dataset (specifically, the nnU-Net training set) to simplify the segmentation task for the network. Since \textit{TotalSegmentator} often produces coarse segmentations that may exclude parts of the colon, the masks were enlarged with 35 voxels dilation to ensure better coverage.\\
To create a dataset for training a fluid segmentation model in \textit{RootPainter}, we randomly selected seven axial slices per scan from regions containing the previously segmented air-filled colon. The scans selection matched the training set for the nnU-Net. The created dataset ensures diverse patient representation and sufficient slices for interactive training. This step resulted in 2,030 slices, which were converted to 1000×1000 PNGs for detailed annotation and segmentation.
We followed the corrective-annotation protocol described in \cite{smith2022rootpainter}, which involves simultaneous annotation and model training. An expert inspected model predictions and assigned corrections, which were then incorporated into the training set to refine subsequent predictions. We evaluated the accuracy of the fluid segmentation model with the Dice score, \textit{i.e.,} the difference between the initial predicted segmentation and the corrected segmentation after user annotation is assigned.\\ 
We applied the resulting model to predict the fluid across the entire dataset, followed by post-processing to refine the results. Components less than 2000 voxels ($\approx 0.002 cm^3$) and less than 2 mm from the surface of the colon were removed. Since CTC is typically performed in prone or supine positions, fluid accumulates below (supine) or above (prone) the colon due to gravity (see Fig. \ref{fig:fluid}). This information was used to refine segmentation by discarding fluid pixels without air-filled colon pixels within ±2 axial slices in the axial plane. Additional post-processing steps included hole filling, Gaussian smoothing for surface refinement, and connecting fluid regions to the nearest air-filled colon by filling empty pixels, in the sagittal plane. The final fluid annotations were validated by an expert, with additional manual post-processing applied to correct errors or fill in missing regions.
%
%
\subsection{Fully automated colon segmentation with nnU-Net}
To develop the fully automated colon segmentation, we utilized the high-resolution colon annotation dataset generated above. The dataset was split into training set (66.6 \%) and test set (33.3 \%), ensuring an even distribution of gender and patient position (supine and prone). The split was not subject-based, meaning the same subject could appear in both training and test sets. This approach was chosen because the colon exhibits high anatomical variability and flexibility, resulting in significantly different shapes when scanned in different positions.\\
We trained four 3D full-resolution nnU-Net v2 \cite{isensee2021nnunet} using two input dataset: raw images and masked images. The masks were produced as described in section \ref{subsec:fluid} from \textit{TotalSegmentator} colon predictions, dilated by 35 voxels. For each of the two datasets, we trained two models. One model that segmented the full colon including air and fluid, and one model that segmented only the air filled region.\\
To refine network predictions, small artifacts were removed by filtering out islands smaller than 2,000 voxels.
The models were evaluated on the test set using three boundary-based metrics: (1) mean absolute surface distance (MASD), (2) average symmetric surface distance (ASSD), and (3) 95th percentile Hausdorff distance (HD 95\%). These metrics, selected based on the \textit{Metrics Reloader} framework \cite{maier2024metrics}, emphasize boundary precision due to the colon tubular shape and high volume-to-surface ratio. We used the \textit{MONAI} implementation of the metrics \cite{MetricsReloaded} and computed them on both raw model predictions and refined predictions (after small island filtering). Since \textit{TotalSegmentator} is the state-of-the-art open-source tool for colon segmentation, we compared our results against its predictions on the test set, evaluating the same metrics for both raw and refined outputs.
\section{Results}
\subsection{Characteristics of the high-resolution annotated colon dataset}
\begin{figure}[tb]
    \centering
    \includegraphics[width=0.99\linewidth]{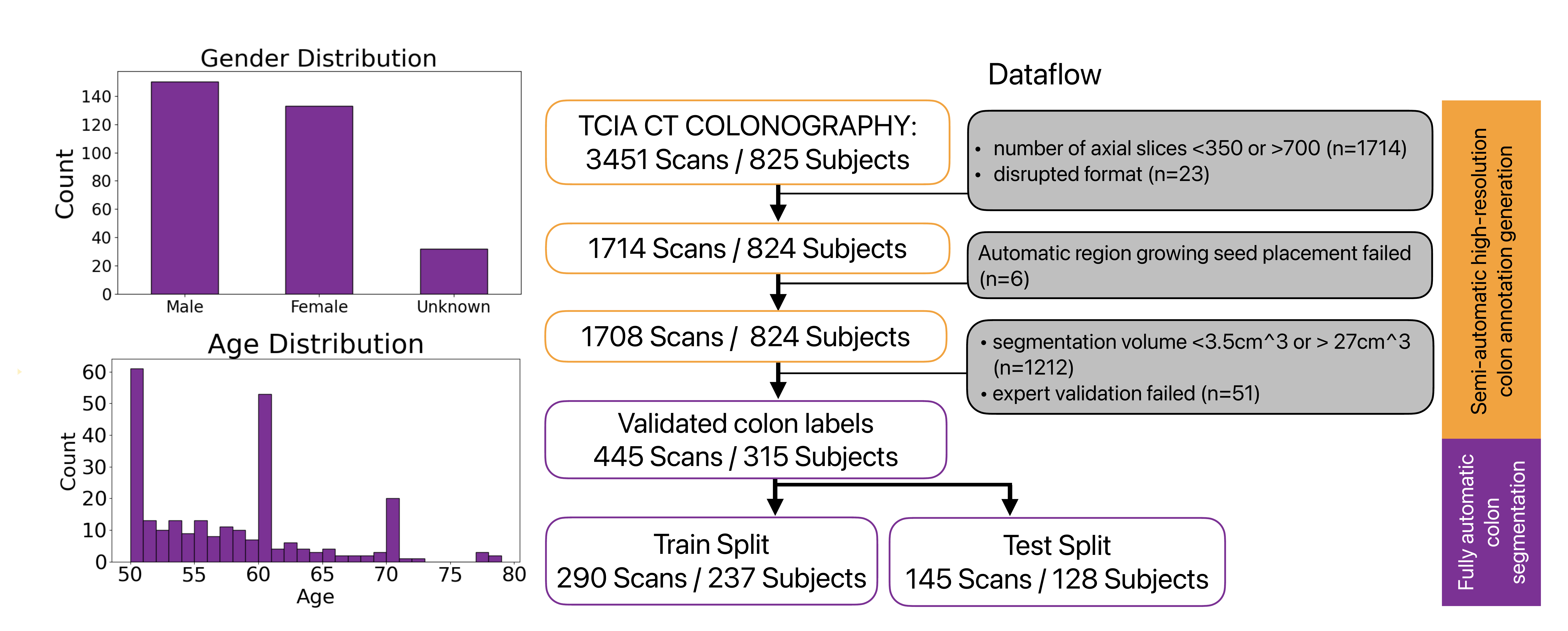}
    \caption{Dataflow for creating the high-resolution annotated dataset to train and test the fully automatic model for colon segmentation. The gray box indicates reasons for scans exclusion. The plots on the left display the gender and age distribution in the final annotated dataset.}
    \label{fig:diagram}
\end{figure}

Of the 3,451 scans in the TCIA CT colonography dataset, 435 were used to train and test an nnU-Net for automatic colon segmentation. Fig. \ref{fig:diagram} illustrates the reasons for exclusion. The final dataset, containing high-resolution validated colon labels, was split into 290 training scans and 145 test scans, stratified by gender and patient position.
\subsection{Annotations of the colon fluid-filled segments}
\begin{figure}[tb]
    \centering \includegraphics[width=1\linewidth]{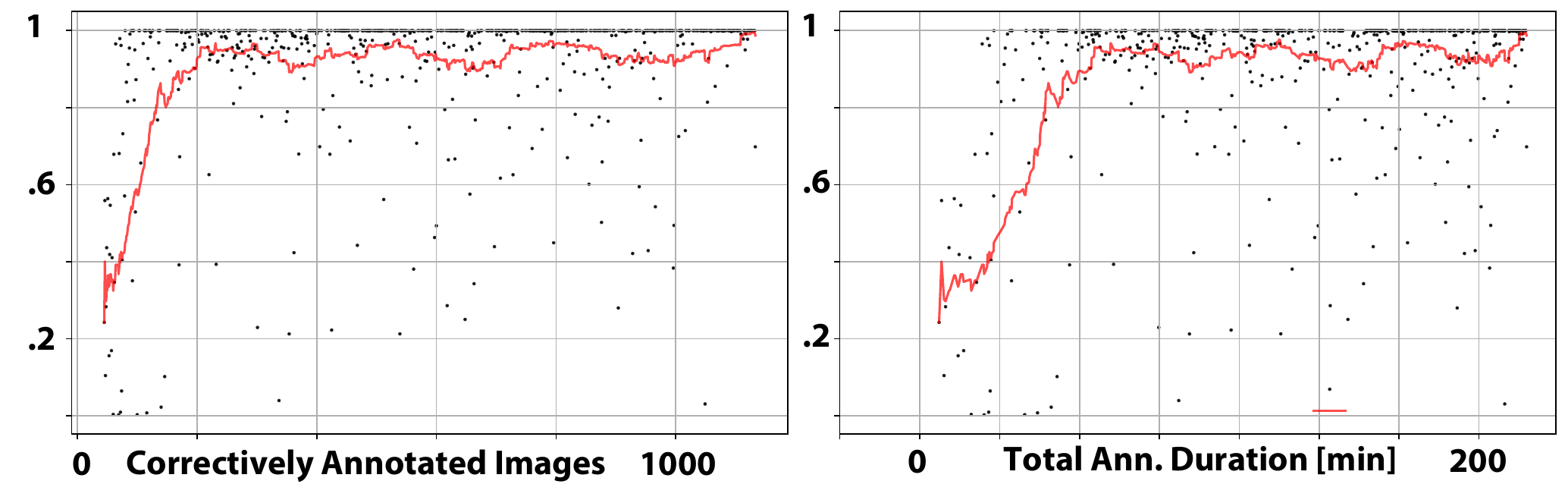}
    \caption{Dice as function of number of annotated images and annotation time.}
    \label{fig:annotdurdicef1}
\end{figure}

The corrective annotation process continued until 1,134 images were evaluated, with annotations assigned to 390 images over 215 minutes. Fig. \ref{fig:annotdurdicef1} illustrates how Dice scores improved and fluctuated over time.
Annotation stopped at image 1,134 when \textit{RootPainter}’s colon segmentation was deemed satisfactory upon visual inspection. The Dice also showed diminishing returns, justifying the decision.
Performance improved most in the first 80 minutes, reaching a rolling Dice above 0.95 by image 210 (Fig. \ref{fig:annotdurdicef1}). The final trained \textit{RootPainter} model was then used to segment the full dataset.
%


\subsection{Evaluation of the fully automatic model}
\begin{figure}[tb]
    \centering    \includegraphics[width=0.99\linewidth]{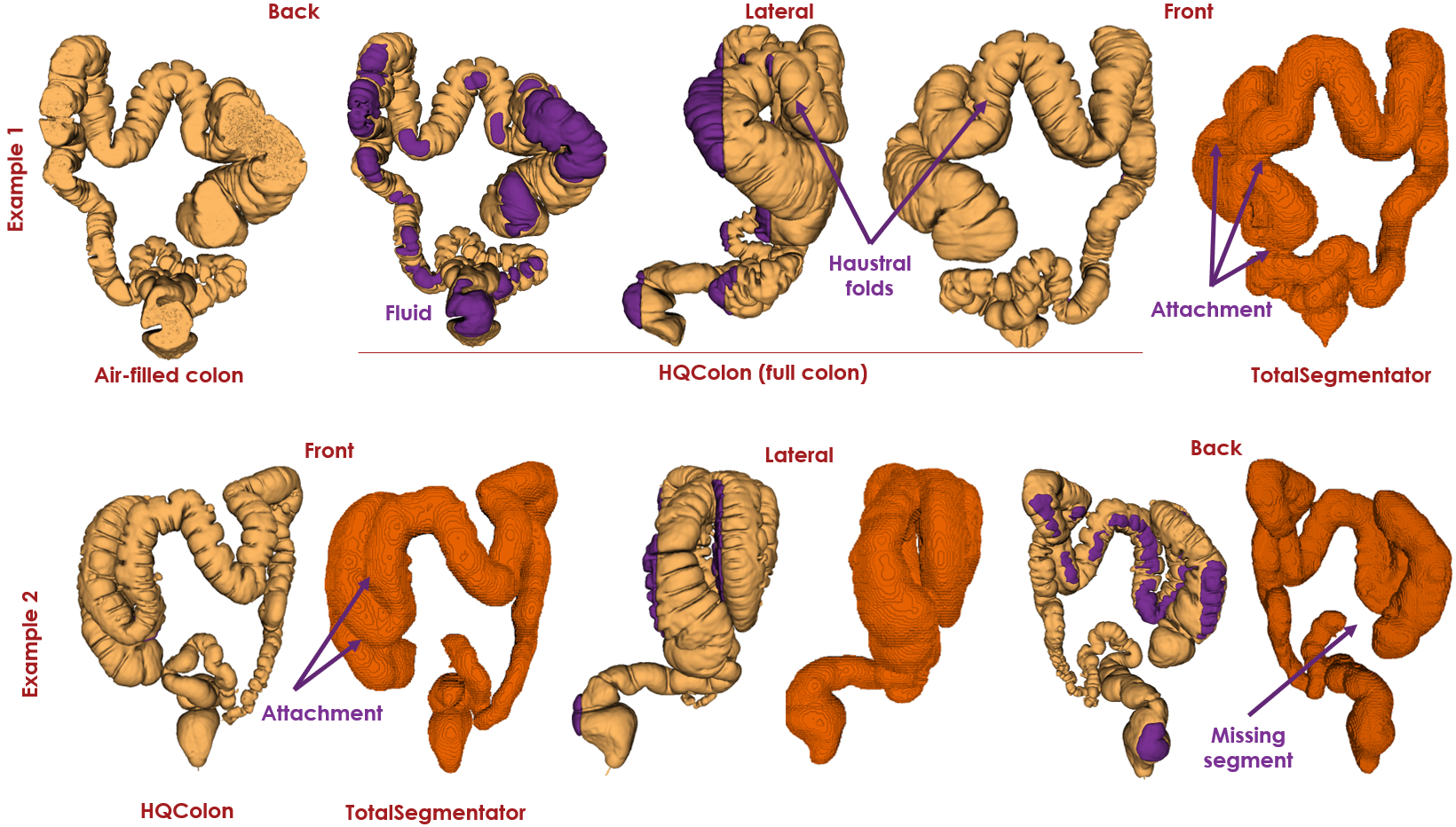}
    \caption{Examples of colon segmentation generated by \textit{HQColon} compared to \textit{TotalSegmentator}. \textit{HQColon} provides high-resolution segmentation, revealing details such as haustral folds, features not captured by \textit{TotalSegmentator}.}
    \label{fig:results}
\end{figure}

\begin{figure}[H]
    \centering
    \begin{subfigure}{0.41\linewidth}
        \centering
        \includegraphics[width=\linewidth]{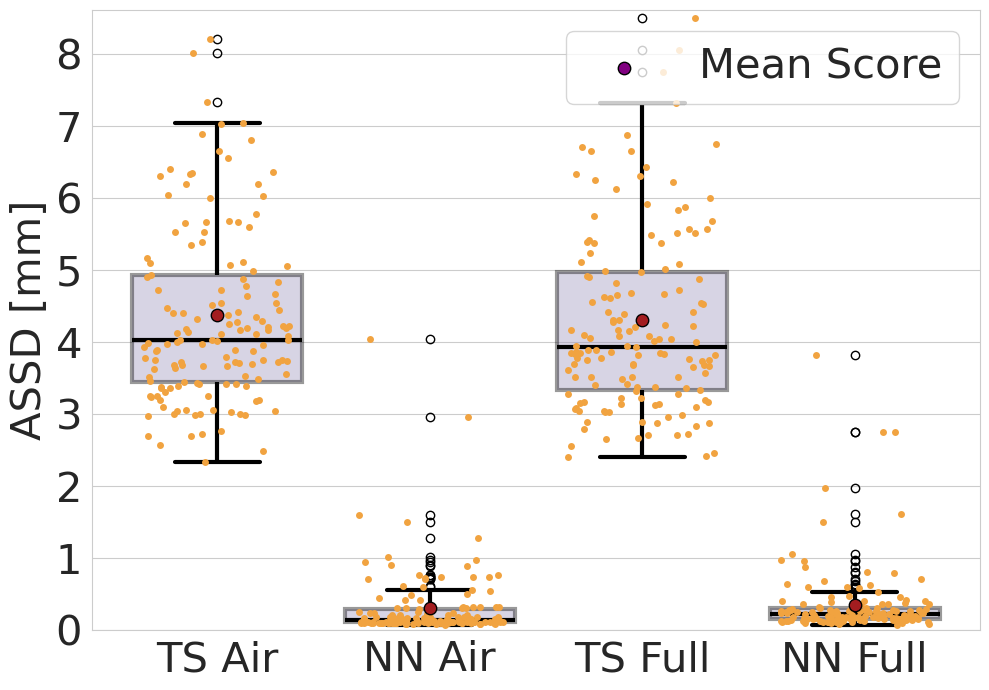}
        \label{fig:ASSD}
    \end{subfigure}
    \begin{subfigure}{0.41\linewidth}
        \centering
        \includegraphics[width=\linewidth]{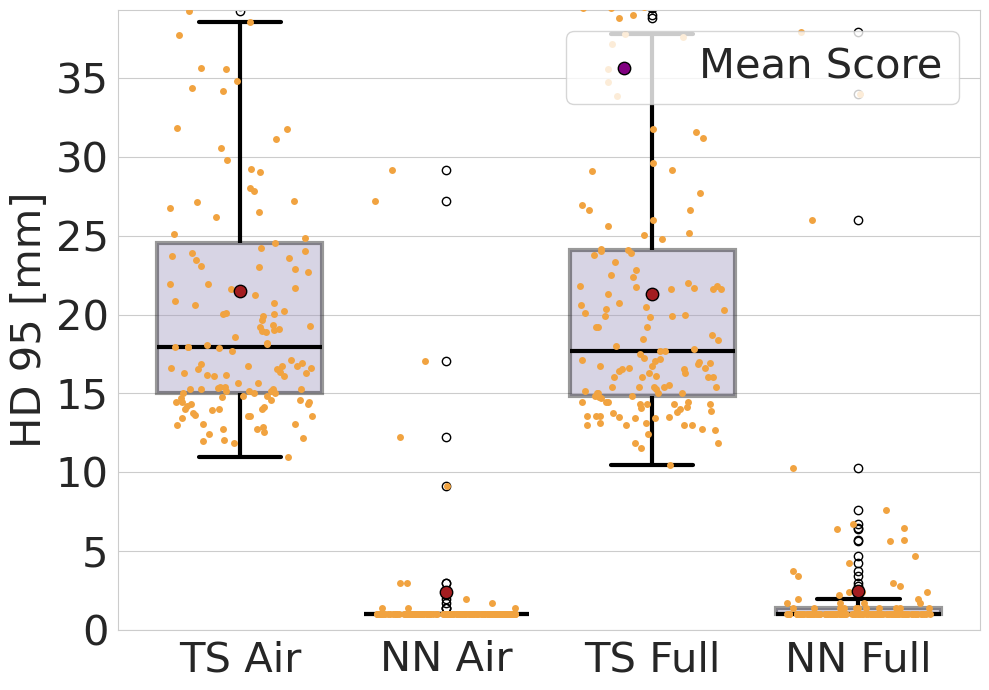}
        \label{fig:95_HD}
    \end{subfigure}
    \caption{Distributions of the Average Symmetric Surface Distance and the Hausdorff 95th percentile for the nnU-Net (NN) using raw image as input predicting air-filled colon segments (Air) or the full colon (Full) vs. \textit{TotalSegmentator} (TS).}
    \label{fig:distributions}
\end{figure}

Table \ref{tab:results} and Fig. \ref{fig:distributions} present the median values of computed metrics across the different trained models, comparing our approach to \textit{TotalSegmentator}. Medians were used due to the non-normal distribution of most metrics. Since ASSD and MASD showed nearly identical results (differences <0.01 mm), only ASSD is reported.\\
Results show a high level of segmentation accuracy across all our models, with no significant differences, regardless of using a mask as input or applying refinement on the prediction. However, compared to \textit{TotalSegmentator}, all models significantly improved performance in both HD 95\% and ASSD. This improvement is also evident in Fig. \ref{fig:results}, where our method consistently delivers high-resolution colon segmentation, including fluid and haustral folds—features not captured by \textit{TotalSegmentator}. Additionally, our approach correctly segments cases where \textit{TotalSegmentator} either misses segments or erroneously merges separate colon segments. One complete high quality colon segmentation with our method, \textit{HQColon}, of a raw CTC image takes approximately 69 sec using an NVIDIA RTX 4090 GPU. 
\begin{table}[tb]
    \centering
    \caption{Medians with 95\% confidence intervals for the Hausdorff 95th percentile and Average Symmetric Surface Distance across different models: nnU-Net (NN) using raw and masked input images (Mask-), and TotalSegmentator (TS), predicting either air-filled colon segments (-Air) or the full colon (-Full), both before and after prediction refinement\\}
    \label{tab:results}   
    \begin{tabular}{l|cc|cc}
        \toprule
        & \multicolumn{2}{c|}{Raw prediction} & \multicolumn{2}{c}{Refined prediction}\\
        \textbf{Method} & \textbf{HD 95\%}& \textbf{ASSD}& \textbf{HD 95\%}& \textbf{ASSD}\\
        \midrule
        NN-Air &  1.00 [1.00, 1.00] & 0.14 [0.12, 0.17] & 1.00 [1.00, 1.00] & 0.13 [0.12, 0.17]\\
        Mask-NN-Air & 1.00 [1.00, 1.00] & 0.12 [0.11, 0.14] & 1.00 [1.00, 1.00]& 0.12 [0.11,0.14]\\       
        TS-Air & 17.97 [16.58,19.65]& 4.03 [3.88,4.23] & 17.97 [16.34,19.24]& 4.03 [3.83,4.29]\\     
        \midrule
        NN-Full & 1.00 [1.00, 1.00] & 0.36 [0.19,0.25]& 1.00 [1.00, 1.00] & 0.22 [0.19,0.25]\\
        Mask-NN-Full & 1.00 [1.00, 1.00] & 0.18 [0.16,0.21] & 1.00 [1.00, 1.00] & 0.18 [0.16,0.22]\\
        TS-Full & 17.72 [16.58,19.90] & 3.94 [3.83,4.18] & 17.72 [16.58,20.10] & 3.94 [3.78,4.18]\\
        \bottomrule
    \end{tabular}
\end{table}
\section{Discussion and Conclusion}
Our results demonstrate that \textit{HQColon} significantly outperforms \textit{TotalSegmentator}, the only other open-source tool for colon segmentation, achieving an order of magnitude higher accuracy (Table \ref{tab:results}). Regardless of input type, segmentation target, or refinement, we observed minimal performance differences in the models we trained (Table \ref{tab:results}). This outcome suggests that nnU-Net can achieve high-accuracy colon segmentation even when trained on raw images without pre-processing or post-processing. This approach eliminates the need for additional mask computation, which would otherwise rely on \textit{TotalSegmentator} and risk missing parts of the colon due to incomplete masks.\\
The fast annotation pipeline we developed addresses a key challenge in medical image analysis: generating high-resolution segmentation models with limited labeled data. Labeling is often time-intensive, even for experts, especially for complex 3D structures like the colon. Our approach reduces user involvement by automatically discarding images where fast annotation fails (\textit{e.g.,} due to incorrect automatic seed placement of the regional growing algorithm). This solution was feasible due to the availability of a large dataset, ensuring that the final subset was sufficient for training nnU-Net. In addition, we leveraged an interactive machine learning approach for the colon fluid segmentation, that allowed us to rapidly annotate, train and review a fluid segmentation model (Fig. \ref{fig:annotdurdicef1}).\\
Our method has been tested only on fully reconstructed colons, excluding collapsed cases. Future work will focus on extending its applicability to collapsed colons to enhance robustness across different clinical scenarios.\\
The overarching goal of this study was to develop a fully automatic, high-resolution colon segmentation method. Our approach, \textit{HQColon}, is the first open-source tool for high-resolution colon segmentation, empowering researchers and clinicians to generate their own segmentations and expand colon annotation datasets. The resulting anatomical models and 3D reconstructions have broad potential applications, including population studies, digital twins, personalized medicine, and AI-driven diagnostic tools.

  \bibliographystyle{splncs04}
  \bibliography{reference}

\end{document}